\documentclass[10pt,twocolumn,letterpaper]{article}

\usepackage{iccv}
\usepackage{times}
\usepackage{epsfig}
\usepackage{graphicx}
\usepackage{amsmath}
\usepackage{amssymb}

\newcommand{\keypoint}[1]{\vspace{0.1cm}\noindent\textbf{#1}\quad}
\usepackage{soul}
\usepackage{arydshln}
\usepackage{pifont}
\usepackage{floatrow}
\usepackage{multicol}
\usepackage{multirow}
\usepackage{rotating}
\newfloatcommand{capbtabbox}{table}[][\FBwidth]

\newcommand{\cmark}{\ding{51}}
\newcommand{\xmark}{\ding{55}}

\usepackage[pagebackref=true,breaklinks=true,letterpaper=true,colorlinks,bookmarks=false]{hyperref}

\iccvfinalcopy 

\ificcvfinal\fi

\begin{document}

\title{SketchLattice: Latticed Representation for Sketch Manipulation}


\author{Yonggang Qi$^{1}$\thanks{Equal contribution} \quad Guoyao Su$^{1*}$ \quad Pinaki Nath Chowdhury$^{2}$ \quad Mingkang Li$^{1}$ \quad Yi-Zhe Song$^{2}$\\
$^{1}$Beijing University of Posts and Telecommunications, CN \quad $^{2}$SketchX, CVSSP, University of Surrey, UK \\
{\tt\small \{qiyg, sgybupt, lmk\}@bupt.edu.cn}  \qquad {\tt\small  \{p.chowdhury, y.song\}@surrey.ac.uk}
}

\maketitle
\ificcvfinal\thispagestyle{empty}\fi

\begin{abstract}
The key challenge in designing a sketch representation lies with handling the abstract and iconic nature of sketches. Existing work predominantly utilizes either, (i) a pixelative format that treats sketches as natural images employing off-the-shelf CNN-based networks, or (ii) an elaborately designed vector format that leverages the structural information of drawing orders using sequential RNN-based methods. While the pixelative format lacks intuitive exploitation of structural cues, sketches in vector format are absent in most cases limiting their practical usage. Hence, in this paper, we propose a lattice structured sketch representation that not only removes the bottleneck of requiring vector data but also preserves the structural cues that vector data provides. Essentially, sketch lattice is a set of points sampled from the pixelative format of the sketch using a lattice graph. We show that our lattice structure is particularly amenable to structural changes that largely benefits sketch abstraction modeling for generation tasks. Our lattice representation could be effectively encoded using a graph model, that uses significantly fewer model parameters {($13.5$ times lesser)} than existing state-of-the-art. Extensive experiments demonstrate the effectiveness of sketch lattice for sketch manipulation, including sketch healing and image-to-sketch synthesis.
\end{abstract}

\vspace{-0.5cm}
\section{Introduction}

\begin{figure}
    \centering
    \includegraphics[width=\linewidth]{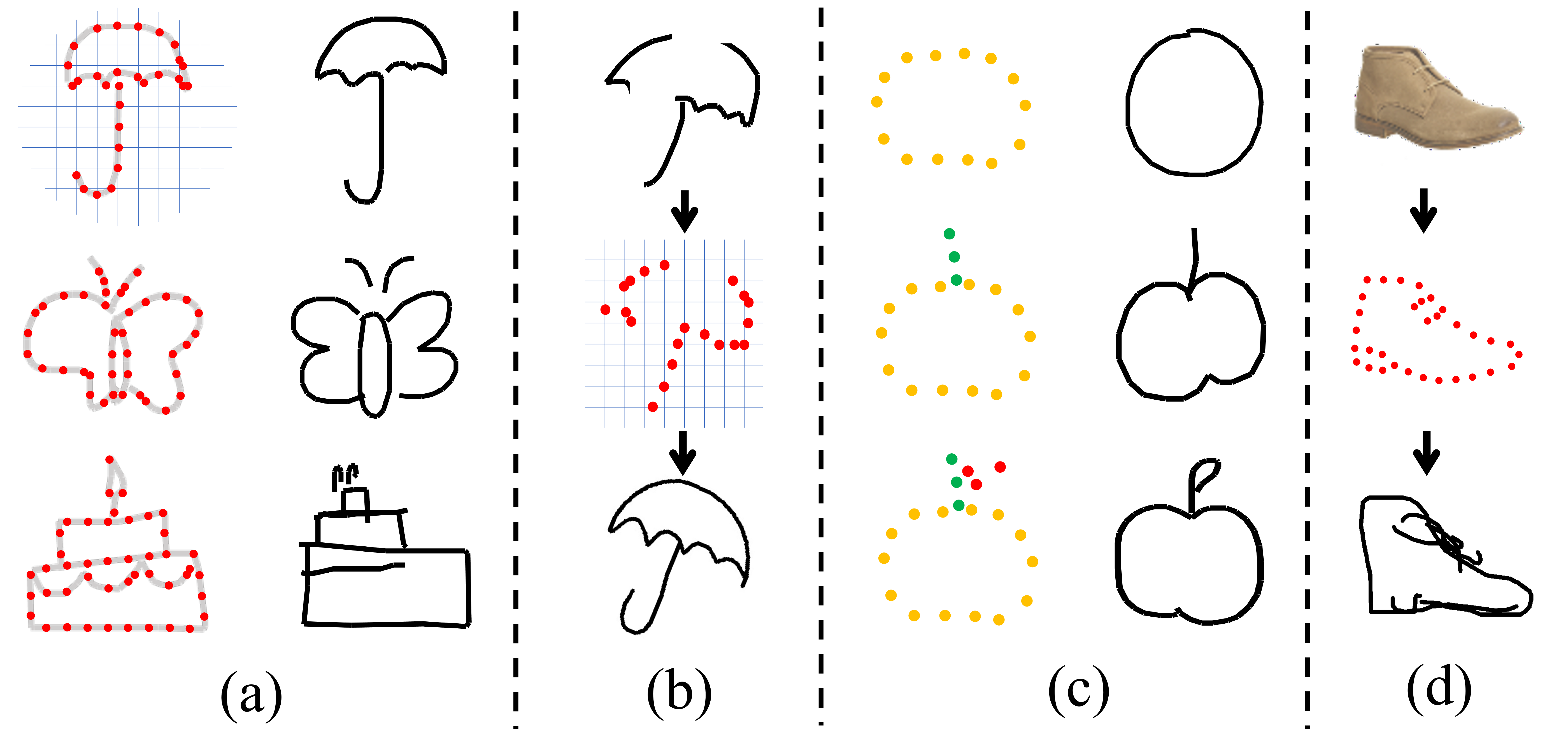}   
    \caption{ (a) Given lattice points sampled on input sketches (Left), our proposed Lattice-GCN-LSTM network can \emph{recreate} a corresponding vector sketch (Right). (b) Given a corrupted sketch, the resulting lattice points are used to reconstruct a \emph{similar} sketch accordingly. (c) The abstraction level of generated sketches is controllable by varying the density of latticed points. (d) Image-to-sketch synthesis by dropping a few lattice points along the edge of an object.}
    \label{fig:intro}
\end{figure}

Research on freehand human sketches has become increasingly popular in recent years. Due to its ubiquitous ability in recording visual objects \cite{eitz2012humans}, sketches form a natural medium for human-computer interaction. Deriving a tailor-made representation for sketches sits at the core of sketch research, and has direct impact on a series of downstream applications such as sketch recognition \cite{yu2015sketch,xu2019multigraph,kipf2016semi}, sketch-based image retrieval \cite{cao2010mindfinder,yu2016sketch,pang2017cross,pang2019generalising}, sketch-3D reconstruction \cite{lun20173d,jiang2019disentangled,wang20203d,ranjan2018generating,shih20203d}, and sketch synthesis \cite{liuunsupervised,song2018learning}. Albeit a pivotal component, designing an effective representation is challenging since sketches are typically abstract and iconic.

Prior works predominantly relied on encoding sketch in a pixelative format (i.e., an image) \cite{eitz2012humans,yu2015sketch,sangkloy2017scribbler}. Although it provided the convenience of using off-the-shelf convolutional neural network effortlessly re-purposed for sketches, pixelative format lacks the intuitive exploitation of structural information. The presence of structural information is vital for sketch abstraction modeling \cite{eitz2012humans,riaz2018learning}, which in turn is essential for downstream tasks that dictate structural manipulation such as sketch generation \cite{ha2017neural,chen2017sketch,karras2019style} and sketch synthesis \cite{liuunsupervised,song2018learning}.

RNN-based approaches have consequently emerged as means to fully explore the sequential nature of sketches \cite{ha2017neural}. The research convention is to use {\em QuickDraw} \cite{quickdraw} vector format where each sketch is represented as a list of offsets in $x$ and $y$. Thanks to the stroke-level modeling, these approaches do offer a degree of flexibility in generation and synthesis tasks, yet they do so by imposing a strong assumption -- all sketches have sequential stroke data to them. This assumption largely prohibits the application of RNN-based methods to work with sketches such as those drawn on a piece of paper. A natural question is therefore -- is there a way to remove the bottleneck on requiring vector data but at the same time preserve the structural cues that vector data provides?

To answer this question, we propose an alternative sketch representation inspired by the concept of lattice structures -- SketchLattice. We define SketchLattice as a set of points sampled from the original 2D sketch image using a lattice graph as shown in Figure~\ref{fig:intro} (a). Such latticed sketch representation, although seemingly simplistic, is remarkably amenable to structural deformations, thereby providing vital benefits for sketch abstraction modeling and in subsequent downstream sketch generation tasks. Our proposed latticed representation can be easily and effectively encoded using a simple off-the-shelf graph convolutional network (GCN) \cite{kipf2016semi,chen2019multi,yang2018graph},  resulting in considerably fewer model parameters ($13.5$ times lesser) as compared to recent state-of-the-art techniques. This not only makes our proposed sketch representation easily deployable, thus making further progress towards practicality, but also reduces the difficulty in optimization and training to give a competitive performance.

Specifically, each point in SketchLattice is regarded as a graph node. Geometric proximity between nodes serves as guiding principle for constructing the adjacency matrix to form graph links. Intuitively, the proposed GCN-based feature extractor learns the topology of points in a sketch object. Despite being simple, our novel sketch representation is surprisingly effective for sketch generation. 
In particular, using the proposed latticed representation, we show how to recover a corrupted sketch using our Lattice-GCN-LSTM network, as represented in Figure~\ref{fig:intro}(b). Additionally, we present a novel aspect in sketch representation, where the abstraction level in the generated sketch is controllable as shown in Figure~\ref{fig:intro}(c), subject to the density of points sampled by the lattice graph. Furthermore, our method is also applicable to the problem of image-to-sketch synthesis by simply dropping a few key points along the edge of a target object as depicted in Figure~\ref{fig:intro}(d).

Our contributions are summarized as follows: (i) we propose SketchLattice, a novel latticed representation for sketches using an extremely simple formulation i.e., a set of points sampled from a sketch image using a lattice graph. (ii) Our latticed representation can be easily and effectively encoded using a simple graph model that use fewer model parameters, thereby making important progress towards efficiency. (iii) We show how the abstraction level of generated sketches is controllable by varying the density of points sampled from an image using our lattice graph.

\section{Related Work}
\keypoint{Sketch Data Format} 
There are mainly two types of data formats for sketch representation -- image-based and sequential representation. While the former treats sketch as a conventional 2D image with pixel values (i.e., the pixelative format), the latter considers sketch as an elaborately designed set of ordered stroke points (i.e., vector format), represented by offset coordinates along $x$ and $y$ directions with pen states (touch, lift and end) \cite{ha2017neural}. Traditional sketch feature extractors are usually CNN-based approaches \cite{yu2015sketch,chen2017sketch} that can directly take sketch image in pixelative format during input. However, this is highly redundant due to the sparsity of line drawings in a sketch image that necessitates heavy engineering efforts \cite{yu2015sketch}. Additionally, CNN-based approaches cannot effectively capture structural cues since it does not encode position and orientation of objects, 
leading to sub-par results on generation models.

In contrast, sequential representation is sketch-specific \cite{quickdraw}, designed according to the drawing habit of humans which is constructed stroke-by-stroke. Such sequential, vector format representation allows modeling sketches using RNN-based methods. This resulted in impressive results such as sketch generation and sketch synthesis using long short-term memory (LSTM) \cite{ha2017neural,su2020sketchhealer}. Although promising, such RNN-based approaches require a vectorized data format at the input, which leads to a major bottleneck limiting practical usage in the absence of vector sketches, like sketches drawn on a piece of paper. Therefore, we aim to propose an unexplored technique, by employing a more practical lattice sketch representation, i.e., SketchLattice, that avoids storing stroke orders while still keeping strong spatial evidence that vector sketches typically provide.

\keypoint{Graphical Sketch Embedding} 
Graph convolutional networks (GCNs) \cite{bruna2013spectral, kipf2016semi} were originally designed to deal with structured data, such as knowledge graphs or social networks, by generalizing neural networks on graphs. In the past few years, exciting developments have been made that explore GCNs capabilities for various vision tasks including image classification \cite{chen2019multi}, captioning \cite{yao2018exploring}, image understanding \cite{aditya2018image}, action recognition \cite{li2019actional}, 3D object detection \cite{zarzar2019pointrgcn}, and shape analysis \cite{wei2020view}. Yet, until recently, a few attempts \cite{yang2020sketchgcn, yang2020s} started to apply GCNs on sketch embedding. The existing visual sparsity and spatial structure of sketch strokes are naturally compatible with graphical representations. However, the dominating approach in sketch research assumes access to a vector format where the stroke orders are required, thus resulting in a major limitation in real-world cases. On contrary, ours provides a generic approach that explores the geometrical proximity in a latticed representation of sketch. We also show how our proposed graphical sketch embedding can be used additionally for tasks, involving sketch generation and image-to-sketch translation.

\section{Methodology}

\begin{figure*}
    \centering
    \includegraphics[width=\linewidth]{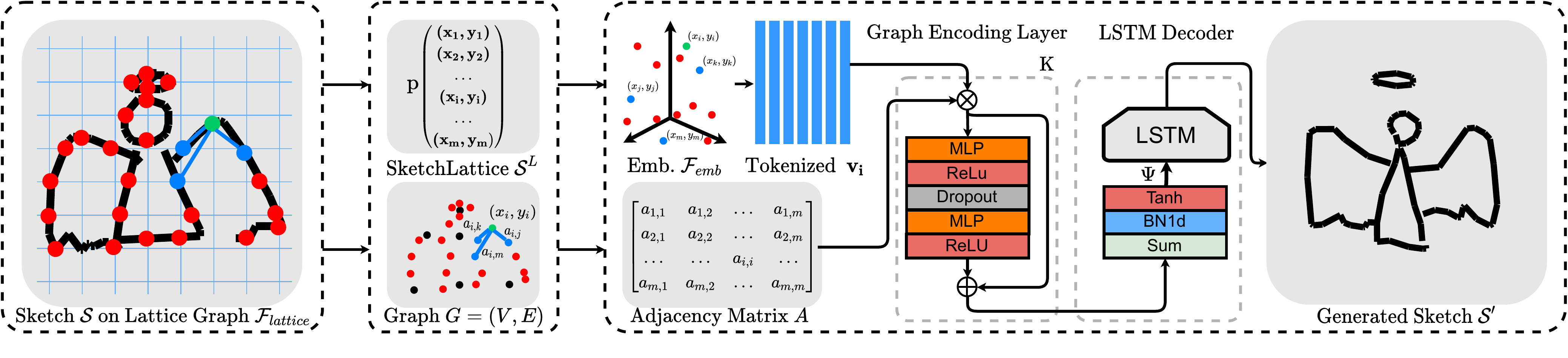}
    \caption{A schematic representation of Lattice-GCN-LSTM architecture. An input sketch image or the edge map of an image object is given to our lattice graph to sample lattice points. All overlapping points between the \emph{dark pixel} in sketch map and uniformly spread lines in lattice graph are sampled. Given the lattice points, we construct a graph using proximity principles. A graph model is used to encode SketchLattice into a latent vector. Finally, a generative LSTM decoder recreates a vector sketch which resembles the original sketch image. }
    \label{fig:lats-gcn-structure}
\end{figure*}

\keypoint{Overview} We describe the Lattice-GCN-LSTM network where the central idea is a novel sketch representation technique that (i) transforms an input 2D sketch image $\mathcal{S}$ into a set of points $\mathcal{S}^{L} = \{p_1, p_2, \dots, p_m \}$, using a lattice graph $\mathcal{F}_{lattice}$. Each point $p_i = (x, y)$ in $\mathcal{S}^{L}$ represents the absolute coordinates $x$ and $y$ in the $\mathcal{S}$. We call $\mathcal{S}^{L}$ as the lattice format representation of $\mathcal{S}$. (ii) Our novel lattice format $\mathcal{S}^{L}$ could be seamlessly transformed to a graphical form $G=(V, E)$ that is encoded into a $d$-dimensional sketch-level embedding vector $\Psi \in \mathbb{R}^{d}$ using a simple off-the-shelf GCN-based model. (iii) We observe how this sketch-level embedding vector $\Psi$ could help in downstream tasks such as sketch generation by using existing LSTM-based decoding models. Figure~\ref{fig:lats-gcn-structure} offers a schematic illustration.

\subsection{Latticed Sketch}\label{sec:latticed-sketch}
The input to our proposed latticed sketch representation is a sketch image $\mathcal{S} \in \mathbb{R}^{w \times h}$ where $w$ and $h$ represent the width and height of $\mathcal{S}$ respectively. We extract the latticed sketch $\mathcal{S}^{L}$ from $\mathcal{S}$ using the lattice graph $\mathcal{F}_{lattice}$. Our lattice graph $\mathcal{F}_{lattice}$ is a grid that constitutes of uniformly distributed $2n$ horizontal and vertical lines, arranged in a criss-cross manner. The optimal value of $n$ for any given sketch image $\mathcal{S}$, could be empirically determined during inference without further training. As shown in Figure~\ref{fig:lats-gcn-structure}, we construct $\mathcal{S}^{L}$ by sampling the set of all overlapping points $\mathcal{S}^{L} = \{p_1, p_2, \dots, p_m\}$ between a \emph{black pixel} in sketch image $\mathcal{S}$ representing a stroke region, and the $2n$ horizontal or vertical lines in $\mathcal{F}_{lattice}$. Formally, we define $\mathcal{S}^{L}$ as:
\begin{equation}
    \mathcal{S}^{L} = \mathcal{F}_{lattice}(\mathcal{S})
\end{equation}

Although extremely simple, this novel latticed {sketch} representation $\mathcal{S}^{L}$ is very informative since it can express the topology (i.e., overall structure and shape) of the original sketch image $\mathcal{S}$, without the need of vector data. Additionally, our latticed sketch representation is very flexible because, (i) there is no constrain for the size of input sketch image $(w \times h)$ and the original aspect ratio is maintained, (ii) the abstraction level of the generated sketches modulates depending on the sampling density from the lattice graph $\mathcal{F}_{lattice}$ by varying the value of $n$. Increasing the value of $n$ would result in more detailed sketches, whereas decreasing it would lead to highly abstract sketches. Figure~\ref{fig:intro}(c) and \ref{fig:abstract} demonstrate how adding more sample points $p_i$ changes the abstraction level of generated sketches.

\subsection{Graph Construction}\label{sec:graph-construction}

\keypoint{Graph Nodes $V$} The SketchLattice $\mathcal{S}^{L}$ can be effectively encoded by a simple graph model which not only consumes fewer model parameters, thereby increasing efficiency, but also allows easier optimization resulting in a model better trained to give a state-of-the-art performance. For each point $p_i \in \mathcal{S}^{L}$ we calculate $\mathbf{v}_i \in V$ representing elements of the set $V$, denoting graph nodes. To ensure that the encoding process is amenable to structural changes, each point $p_i$ is tokenized by a learnable embedding function $\mathcal{F}_{emb}(\cdot): \mathbb{R}^{2} \mapsto \mathbb{R}^{d}$ that maps the absolute point location $p_i=(x,y)$ to a $d$-dimensional vector space. Formally,
\begin{equation}
    \mathbf{v}_i = \mathcal{F}_{emb}(p_i)
\end{equation}
where $\mathbf{v}_i \in \mathbb{R}^{d}$ is the resulting tokenized vector representation. Maintaining the original aspect ratio, we resize and pad the input sketch image $\mathcal{S} \in \mathbb{R}^{w \times h}$ to a size of (256, 256) before applying the lattice graph $\mathcal{F}_{lattice}$. Hence, the vocabulary size of the learned embedding function $\mathcal{F}_{emb}$ is $256^2$. Our intuition is that, by using an embedding function $\mathcal{F}_{emb}$ that tokenizes each point location $(x, y)$ to a $d$-dimensional vector $\mathbf{v}_i$, the model would learn to obtain similar embedding features for nearby points. Hence the resulting representation would be more robust to the frequently observed shape deformation and be more amenable that largely benefits sketch abstraction modeling in the generation tasks.

\keypoint{Graph Edges $E$} A straight-forward yet efficacious approach based on the geometric proximity principles is adopted to construct the graph edge links among nodes, based on the corresponding latticed points' locations $p_i=(x,y)$. 
Specifically, we first compute the euclidean distance between every pair of nodes $(\mathbf{v}_i, \mathbf{v}_j)$ by $d_{i,j} = ||p_i - p_j||_{2}$. Then we follow either of the two options: (i) Each node $\mathbf{v}_i \in V$ is connected to its \emph{nearest} neighbor or (ii) each node $\mathbf{v}_i \in V$ is connected to its \emph{nearby} neighbors that are \emph{``close enough''}, i.e., $norm(d_{i,j})<d_{T}$, where $norm(d_{i,j})$ is a normalized distance in (0,1). $d_{T}$ is a pre-defined distance threshold whose value is empirically found to be $0.2$ in our case. An adjacency matrix $\mathbf{A} \in \mathbb{R}^{m \times m}$ is constructed by setting the link strength to $a_{i,j} = 1 - norm(d_{i,j})$ for a pair of linked nodes ($v_i, v_j$), such that a smaller distance would result in a larger score. All the disconnected nodes, $a_{i,j}$ are set to $0$.

\subsection{Graphical Sketch Encoder}
Given the graph nodes $V=\{\mathbf{v}_1, \mathbf{v}_2, \dots, \mathbf{v}_m\}$ and their corresponding adjacency matrix $\mathbf{A}$, we employ a simple graph model to compute our final sketch-level latent vector $\Psi \in \mathbb{R}^{d}$. The resulting vector $\Psi$ allows downstream applications including sketch healing, and image-to-sketch translation. We employ a stack of $K$ identical graph encoding layers, followed by a fully connected (FC) layer, batch normalization and non-linear activation function $Tanh$.

For each $i^{th}$ node $\mathbf{v}_i^k$ in graph encoding layer $k \in [1, K]$, a feature propagation step is executed to produce the updated node feature $\hat{\mathbf{v}}_i^{k}$, where each node $\mathbf{v}_i^k$ attends to all its linked neighbors with non-zero link strength, defined in the adjacency matrix $\mathbf{A}$. We compute $\hat{\mathbf{v}}_i^k$ as:
\begin{equation}
    \hat{\mathbf{v}}_i^{k} = \sum_{j=1}^{m} a_{i, j} \mathbf{v}_j^{k}
\end{equation}
Such a mechanism incorporating spatial awareness not only facilitates message passing among connected nodes, but also adds robustness to missing parts in a lattice sketch while encoding. This greatly benefits downstream tasks such as sketch healing \cite{su2020sketchhealer}. A graph convolution is applied to the resulting rich spatially dependent feature $\hat{v}_i^k$ as:
\begin{equation}
    \mathbf{v}_i^{k+1} = [ReLU(MLP_{\Theta}(\hat{\mathbf{v}}_i^k))]_{\times 2}
\end{equation}
where each encoding layer consists of two multi-layer perceptron (MLP) units, both of which is followed by a rectified linear unit (ReLU). We employ dropout and residual connection in each encoding layer as shown in Figure~\ref{fig:lats-gcn-structure}. The final feature vectors of nodes from the $K^{th}$ graph encoding layer are integrated into a single vector which is further fed into a sequence of FC layer, batch normalization, and $Tanh$ to compute our sketch-level latent representation $\Psi \in \mathbb{R}^{d}$.

\subsection{Sketch Generation by LSTM decoder}
\label{lstm}
Following \cite{ha2017neural,chen2017sketch,su2020sketchhealer}, we design a generative LSTM decoder that generates the sequential sketch strokes in vector format. Accordingly, the sketch-level latent vector $\Psi$ is projected into two vectors $\mu \in \mathbb{R}^{d}$ and $\sigma \in \mathbb{R}^{d}$, then from which we can sample a random vector $z \in \mathbb{R}^{d}$ by using the reparameterization trick \cite{kingma2014vae} to introduce stochasticity in the generation process via an IID Gaussian variable $\mathcal{N}(0,I)$:
\begin{equation}
\label{z}
\begin{aligned}
& z=\mu + \sigma \odot \mathcal{N}(0, I) \\
& \hspace{-0.5cm} \mu=W_\mu \Psi, \; \; \; 
\sigma = exp(\cfrac{W_\sigma \Psi}{2})
\end{aligned}
\end{equation}
where $W_{\mu}$ and $W_{\sigma}$ are learned through backpropagation \cite{ha2017neural}. The latent vector $z$ is used as a condition for the LSTM decoder to sequentially predict sketch strokes. Specifically, the output stroke representation $s_{t-1}$ from the previous time step, together with latent vector $z$ serve as inputs to update the LSTM hidden state $h_{t-1}$ by:  
\begin{equation}
    h_{t} = LSTM_{forward}(h_{t-1}; [s_{t-1}, z])
\end{equation}
where $[\cdot]$ represents concatenation operation. Next, a linear layer is used to predict a output stroke representation for current time step, i.e., $s_{t} = W_s h_t + b_s$, where $W_s$ and $b_s$ are learnable weight and bias. The final stroke coordinates are derived from $s_t$ with the help of Gaussian mixture models, to generate the vector sketch format, represented by $\mathcal{S'}$. We refer readers to \cite{ha2017neural, quickdraw} for more details.

\subsection{Model Training and Deployment}

Our proposed graphical sketch encoder and the generative LSTM decoder are trained end-to-end for sketch generation. Note that, although we require vector sketches to train the LSTM decoder for the purpose of vector sketch generation, our model fully works on image sketch input, rather than vector data during inference. Following \cite{ha2017neural}, the goal is to minimize the negative log-likelihood of the generated probability distribution to explain the training data $\mathcal{S}$, which can be defined as:
\begin{equation}
\min E_{q_{\phi}(z|\mathcal{S})}[- \log p_{\theta}(\mathcal{S}|z)]
\end{equation}
which seeks to reconstruct the vector sketch representation $\mathcal{S}$ from the predicted latent vector $z$. Upon training, the decoder generates a vector sketch conditioned on the graphical encoded latent vector $z$, obtained from our lattice sketch $\mathcal{S}^{L}$ given any image sketch, thus being more effective for practical applications.

\section{Experiments}
The ability to be amenable to structural changes and enable appropriate abstraction modeling for sketch generation are the two key aspects that our proposed SketchLattice representation aims to address. Specifically, we adopt the challenging task of sketch healing to testify the robustness of our novel sketch representation towards frequently occurring structural deformation in sketches.
Additionally, we also observe how our proposed approach could be utilized to perform the task of image-to-sketch translation. 

\keypoint{Implementation details} We implement our model on PyTorch \cite{paszke2017automatic} using a single Nvidia Tesla T4 GPU. Optimization is performed using the Adam optimizer with parameters $\beta_{1}=0.9$, $\beta_{2}=0.99$ and $\epsilon=10^{-8}$. Value of learning rate is set to $10^{-3}$ along with a decay rate of $0.999$ in every iteration. A gradient clipping strategy is adopted to prevent gradient from exploding during the training of LSTM decoder. Essentially, we force the gradient value to $1.0$ if the actual value is larger than $1.0$. The optimal value for the number of graph encoding layer is $K=2$.

\subsection{Sketch Healing}
The task of sketch healing \cite{su2020sketchhealer}  was proposed akin to vector sketch synthesis. Specifically, given a partial sketch drawing, the objective is to recreate a sketch which can best resemble the partial sketch.

\keypoint{From full $\mathcal{S}^{L}$ to partial $\hat{\mathcal{S}^{L}}$} Given the lattice sketch representation $\mathcal{S}^{L}$ from an input sketch image $\mathcal{S}$, we randomly drop a fraction of lattice points in $\mathcal{S}^{L}$ with some probability $P_{mask}$ to generate a partial SketchLattice, represented by $\hat{\mathcal{S}^{L}}$. Hence, the graph edges linked to the removed node are also disconnected, thereby simultaneously modifying the adjacency matrix $\mathbf{A}$. One can think of $P_{mask}$ as the corruption level of an input sketch image.

\subsubsection{Experimental Settings} 
\label{sec:es}
\keypoint{Dataset} Following the footsteps of \cite{xu2019multigraph,su2020sketchhealer}, we use \emph{QuickDraw} \cite{quickdraw} for evaluation since it is currently the largest doodle sketch dataset. More specifically, a small subset of $10$ categories are selected such that it includes (i) both complex and simple drawings, (ii) intra-category objects having high degree of resemblance among each other, and (iii) presence of common life object categories that contain diverse sub-categories such as \texttt{bus} and \texttt{umbrella}. In each category we use $70k$ training and $1k$ testing sketches. The selected $10$ categories are as follows: \texttt{airplane}, \texttt{angel}, \texttt{apple}, \texttt{butterfly}, \texttt{bus}, \texttt{cake}, \texttt{fish}, \texttt{spider}, \texttt{The Great Wall}, \texttt{umbrella}.

\keypoint{Competitors} We compare our proposed Lattice-GCN-LSTM network with three most popular alternatives for vector sketch generation: \textbf{SketchRNN} (SR) \cite{ha2017neural}, \textbf{SketchPix2seq} (Sp2s) \cite{chen2017sketch}, and \textbf{SketchHealer} (SH) \cite{su2020sketchhealer}. Input to SketchRNN is a set of offsets in $x$ and $y$ directions from a vector sketch representation. The key to SketchRNN is a sequence-to-sequence model which is trained without the KL-divergence term. This is done to maintain the fairness of comparison since the KL-divergence term has shown to be beneficial for multi-class scenarios \cite{chen2017sketch}. SketchPix2seq on the other hand replaces its encoding module with a CNN-based encoder that accepts a pixelative format i.e., sketch image. It is expected that such a design will help capture better visual information. Note that, although both SketchRNN and SketchPix2seq were not specifically designed for the task of sketch healing, following \cite{su2020sketchhealer}, we employ these techniques since they are procedural-wise compatible after being re-purposed. The only work specifically addressing the task of sketch healing is SketchHealer \cite{su2020sketchhealer}. Given a vector sketch as input, SketchHealer converts it into a graphical form where each stroke is considered as a node. Next, visual image patches are extracted from each node region. A GCN-based model is applied for encoding a random vector $z$. The decoding procedure of SketchHealer is identical to ours where $z$ is fed into a generative LSTM decoder to generate the corresponding vector sketch. Additionally, we retrain a variant of SketchHealer only using visual cues ({\bf SH-VC}), for which stroke order is unavailable. Hence, geometric proximity principles are utilized to construct graph edges, similar to ours.  
This examines the performance of SH \cite{su2020sketchhealer} in the absence of vector sketches.

\keypoint{Evaluation setup} We adopt a similar evaluation setup in \cite{song2018learning} and \cite{su2020sketchhealer} for quantitative evaluation to understand the effectiveness of our novel latticed sketch representation. First, we evaluate the quality of generated vector sketches (transformed to pixelative format), via \emph{sketch recognition accuracy}. A pre-trained multi-category classifier with AlexNet architecture is used, which is trained on the training split of $345$ \emph{QuickDraw} categories. Higher recognition accuracy essentially signifies the ability of the network to generate realistic  sketches. It also indicates that the network is able to accurately model the underlying data distribution by effectively encoding a sketch into an accurate and informative sketch representation. We use $1000$ testing sketches from each of the $10$ selected categories for a thorough evaluation. Second, we judge the recognizability of the encoded sketch-level latent vector $\Psi$ by performing a \emph{sketch-to-sketch retrieval task}. The objective is, given the encoded representation of a sketch $\Psi$, we expect to retrieve sketches of the same category from a gallery of sketches.  A higher retrieval accuracy signifies that the network has a strong \emph{sketch healing} ability, due to its amenable and robust sketch representation.

\begin{figure*}
    \centering
    \includegraphics[width=\linewidth]{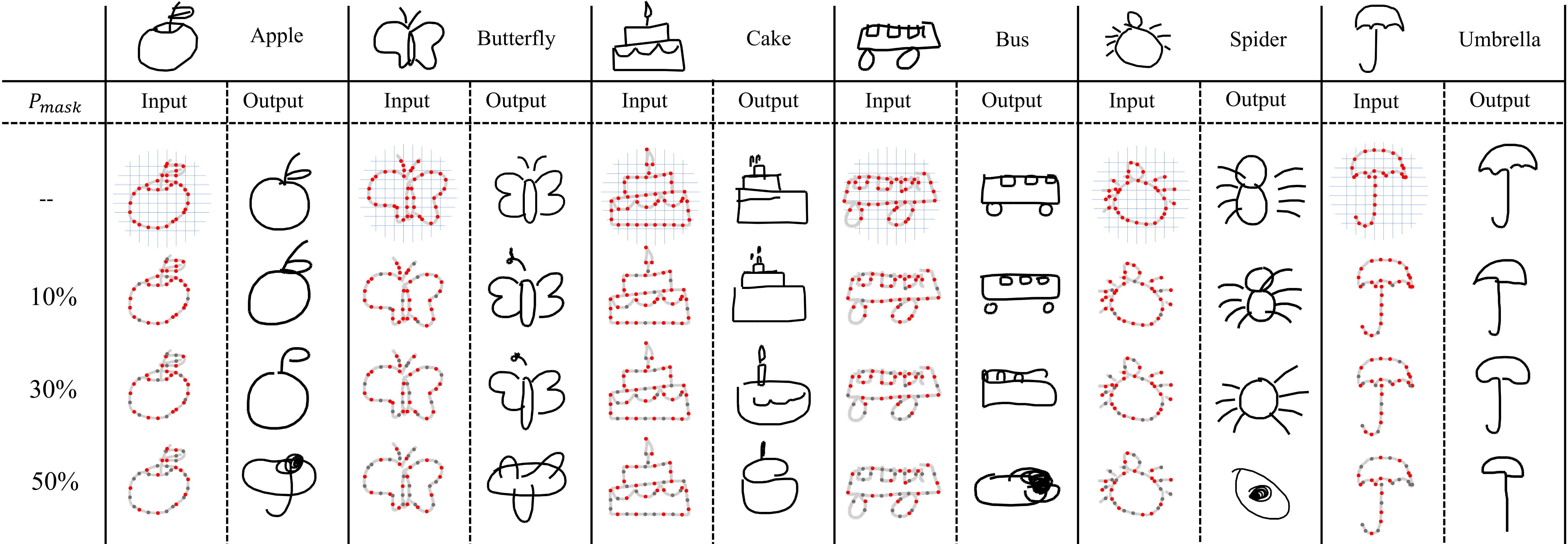}
    \caption{Exemplary results of generated sketch from SketchLattice under different corruption level of mask probability $P_{mask}$ in \emph{QuickDraw} dataset. With an increase of $P_{mask}$, the generated sketch becomes more abstract. For $P_{mask}\leq30\%$ we observe satisfactory generated sketches, but for $P_{mask}=50\%$, the generated new sketches are struggle to faithfully recover the original sketch.}
    \label{fig:mask}
\end{figure*}

\begin{figure}

    \centering
    \includegraphics[width=\linewidth]{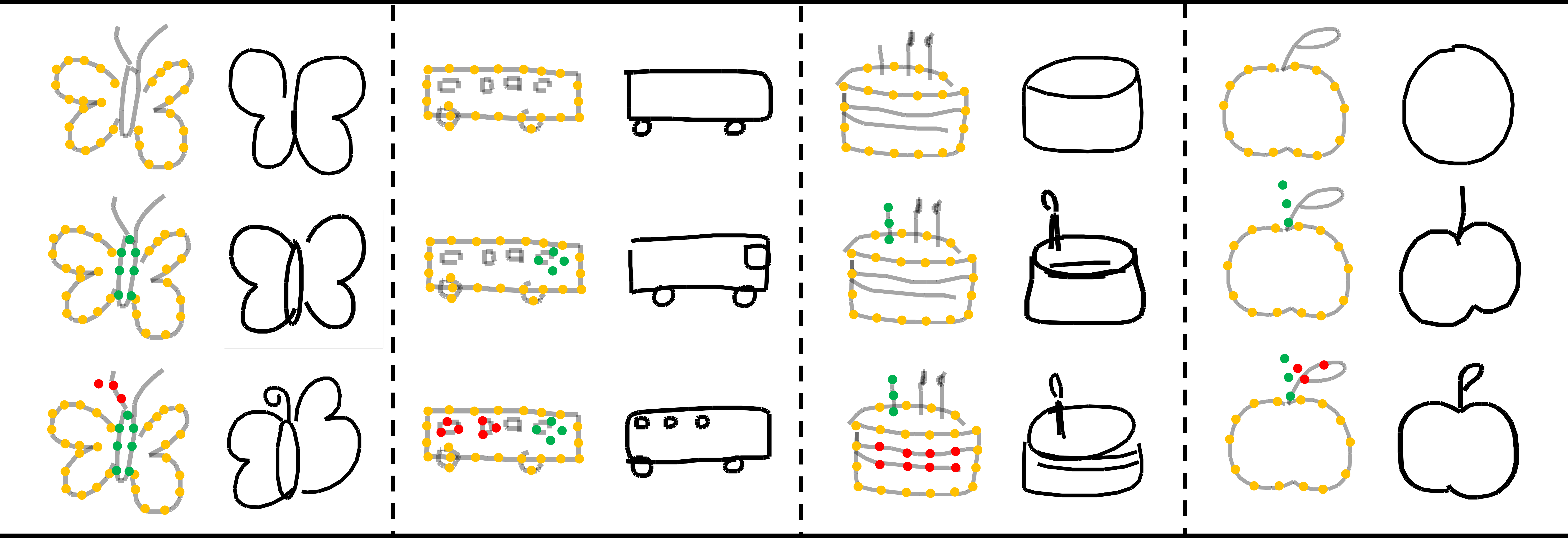}
    \caption{Examples showing how adding more lattice points in different stages (color coded), result in the Lattice-GCN-LSTM network to progressively generate more detailed representation of the category.}
    \label{fig:abstract}
  
\end{figure}

\subsubsection{Results} 

\keypoint{Qualitative Results} We illustrate some examples produced by our lattice-based sketch generator under different values of $P_{mask}$ in Figure~\ref{fig:mask}. We can observe that (i) our latticed representation is robust to partial missing parts such that ours can still generate a novel complete sketch even up to $P_{mask} = 30\%$. (ii) The generated sketch is sensitive to the number of the obtained sampling points, where more points lead to greater details in the generated sketch. Taking the example of \texttt{cake} as shown in Figure~\ref{fig:mask}, we observe how the bottom and candle regions are simplified when increasing $P_{mask}$. (iii) On further lifting $P_{mask}$ to $50\%$, our model can hardly generate a satisfactory sketch, given the severe absence of input sampling points.

We further show that the abstraction level of generated sketch modulates when varying the number and position of input lattice points, as shown in Figure~\ref{fig:abstract}. For example, if we only drop points to form the wings of butterfly, the resulting generated butterfly will be extremely simple. While the body and antenna start to show up when we produce more points to convey the intention of such corresponding details. Similar trends can be found in other classes as well.

\setlength{\tabcolsep}{2pt}
\begin{table}[!t]
    \centering
    \caption{Recognition accuracy (Acc) and Retrieval results (Top-1) according to different corruption levels ($P_{mask}$).  We use ``nearby'' proximity principle and optimal value of $n=32$. Notice that neither vector format (VF) input nor visual cues (VC) are required by using our method during test. }
    
\begin{tabular}{ccccccc}
\hline
Method & VF & VC & \#Params & $P_{mask}$ & Acc & Top-1  \\
\hline\hline
 \multirow{2}{*}{SR \cite{ha2017neural}} & \multirow{2}{*}{\cmark} & \multirow{2}{*}{\xmark} & \multirow{2}{*}{0.67 M} & 10\% & 25.08\% &  50.65\% \\
 & & & & 30\% & 3.44\%  &  43.48\% \\
\hdashline
\multirow{2}{*}{Sp2s \cite{chen2017sketch}} & \multirow{2}{*}{\xmark} & \multirow{2}{*}{\cmark} & \multirow{2}{*}{1.36 M} & 10\% & 24.26\%  &  45.20\% \\
 & & & & 30\% & 10.54\%  &  27.66\% \\
\hdashline
 \multirow{2}{*}{SH \cite{su2020sketchhealer}} & \multirow{2}{*}{\cmark} & \multirow{2}{*}{\cmark} & \multirow{2}{*}{1.10 M} & 10\% & {\color{blue}}50.78\% &  {\color{red}} 85.74\%\\
 & & & & 30\% & {\color{blue}}43.26\% & {\color{red}} 85.47\%  \\
 \hdashline
 \multirow{2}{*}{SH-VC \cite{su2020sketchhealer}} & \multirow{2}{*}{\xmark} & \multirow{2}{*}{\cmark} & \multirow{2}{*}{1.10 M} & 10\% & - &  58.48\%\\
 & & & & 30\% & - & 50.87\%  \\
\hdashline
 \multirow{2}{*}{Ours} & \multirow{2}{*}{\xmark} & \multirow{2}{*}{\xmark} & \multirow{2}{*}{0.08 M} & 10\% & {\color{red}}55.50\% & {\color{blue}}76.02\%  \\
 & & & & 30\%& {\color{red}}54.79\% & {\color{blue}}73.71\%  \\
\hline
\end{tabular}
    \label{tab:recognition}
    \vspace{-0.5cm}
\end{table}

\begin{figure}
    \centering
    \includegraphics[width=\linewidth]{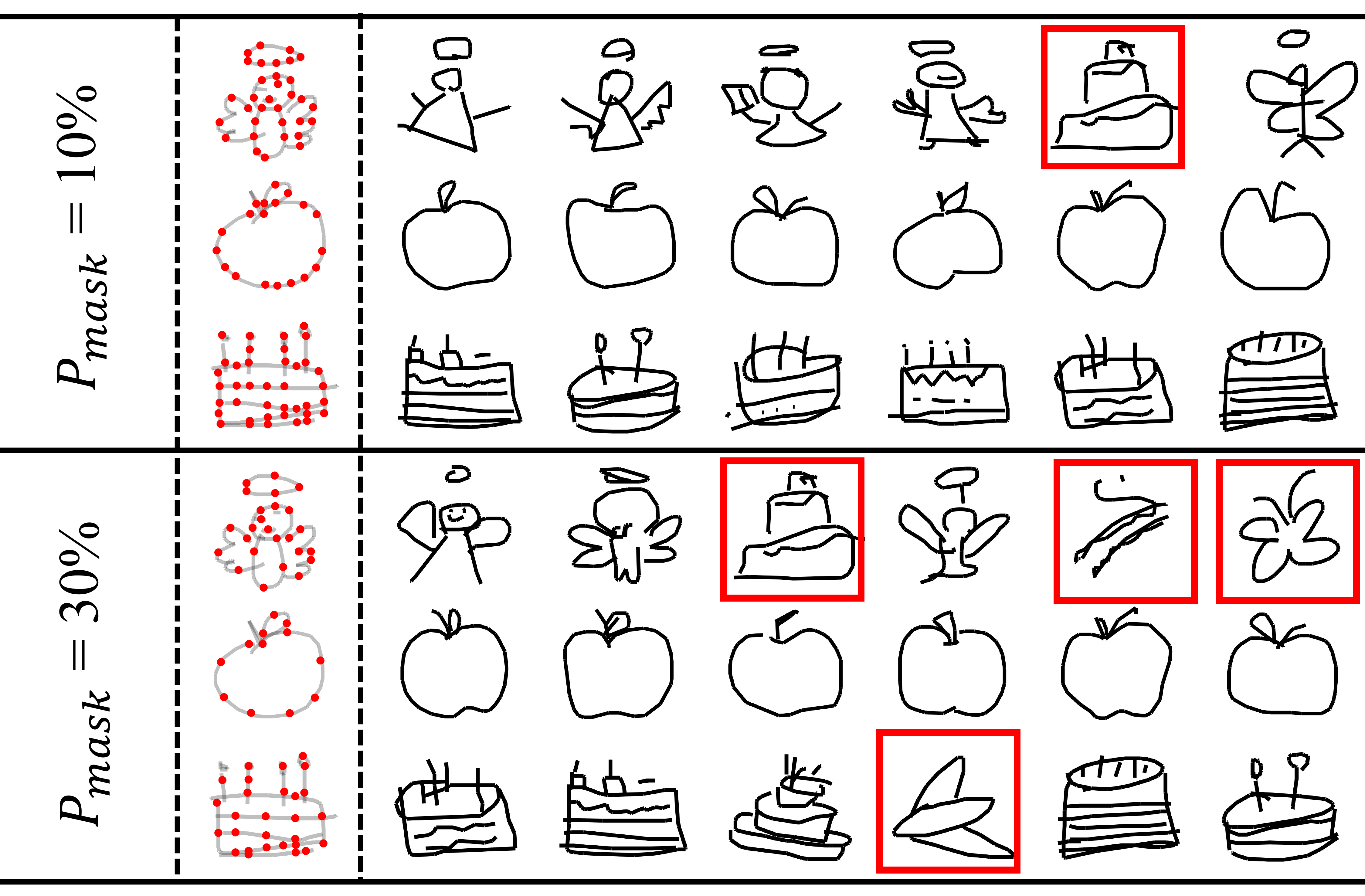}
    \caption{Qualitative retrieval results (Top 6) for sketch-to-sketch retrieval for \emph{QuickDraw} categories under different corruption levels of $P_{mask}$. Red bounding boxes denote false positive.}
    \label{fig:retrieval}
\end{figure}

\keypoint{Quantitative Results} As discussed in Section~\ref{sec:es}, we compare the performance of different models under the two metrics (recognition accuracy and Top-1 retrieval) as shown in Table~\ref{tab:recognition}. We can observe from Table~\ref{tab:recognition} that our approach outperforms other baseline methods on recognition accuracy, suggesting that the healed sketches obtained from ours are more likely to be recognized as objects in the correct categories. Importantly, we can also observe that, unlike other competitors which are very sensitive to the corruption level, ours can maintain a stable recognition accuracy even when $P_{mask}$ increases up to $30\%$. For the task of sketch-to-sketch retrieval, we can see that ours achieves the second best, which is inferior to SketchHealer. However, SketchHealer depends heavily on stroke order provided by vectorized sketches, evidenced by the dramatic decrease of retrieval performance (Table~\ref{tab:recognition} SH vs SH-VC). This signifies the importance and superiority of our approach in the absence of vector input, which is a common practice in real-world cases. In addition, our network is much lighter than the other competitors having far less parameters ($13.5$ times lesser than SketchHealer), since our approach avoids the use of expensive CNN-based operation. We further include two classes \texttt{circle} and \texttt{clock} as distraction to \texttt{apple} for evaluation. A slightly better result 77.80\% (vs 76.02\%) can be observed (12 classes, $P_{mask}$=10\%), suggesting good scalibility and robustness of our latticed representation. Figure~\ref{fig:retrieval} shows some examples of sketch-to-sketch retrieval.

\begin{figure}
    \centering
    \includegraphics[width=\linewidth]{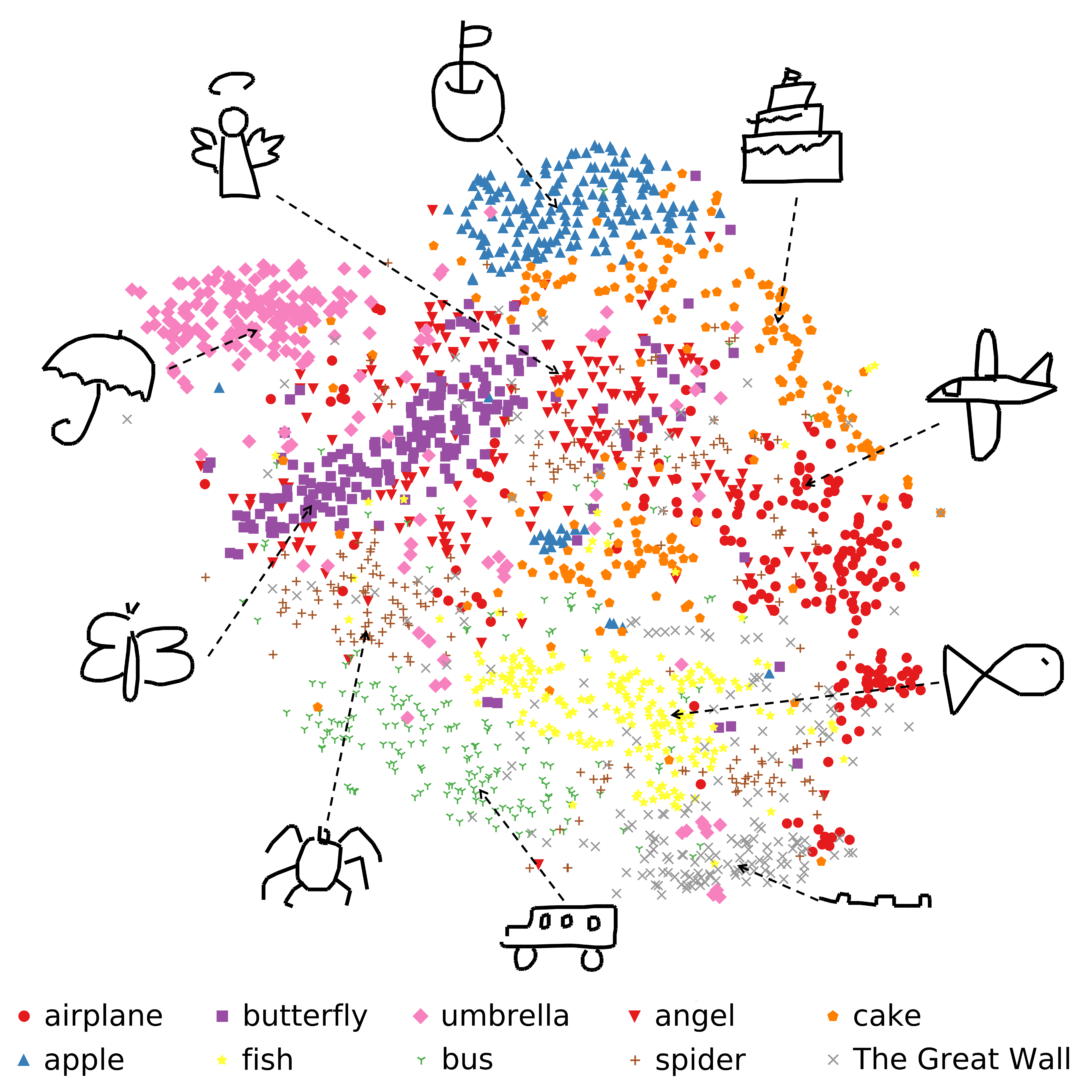}
    \caption{T-SNE plot on sketch-level latent vector $\Psi$ for the selected $10$ \emph{QuickDraw} categories, that demonstrate the discriminative power of our lattice-based graphical encoder. Intra-category instances tend to cluster together, suggesting category-level discriminative ability.}
    \label{fig:tsne}
\end{figure}

\keypoint{Visualization of $\Psi$}
To further visualize the discriminative power of our lattice-based graphical encoder, we randomly select $100$ sketches from each class in the test set, and visualize their latent vectors $\Psi$ using t-SNE in Figure~\ref{fig:tsne}. We observe that intra-category instances tend to cluster together, suggesting a category-level discriminative ability.

\keypoint{Ablation Study} A thorough ablative study using sketch recognition accuracy, is conducted to verify the effectiveness of our different design choices, such as (i) the size of lattice graph, (ii) proximity principle for graph construction, and (iii) importance of residual connection used in our graphical sketch encoder. As shown in Table~\ref{tab:abla}, we can observe that (i) increasing the value of $n$, that will simultaneously increase density of sampling lattice points from $\mathcal{S}^{L}$, directly transpires in an improvement of recognition accuracy. From $n>32$ we start to observe saturation of performance, thereby indicating the optimal value for $n=32$. (ii) For construction of graph, we observe that more neighbors (nearby) works better than using only the nearest one for graph construction. (iii) removing residual connection leads to significant drop in performance, thereby establishing its importance.

\setlength{\tabcolsep}{6pt}
\begin{table}[t]
\small
  \centering
  \caption{Ablative study (Top-1 and Top-3 recognition accuracy) on \emph{QuickDraw} measuring the contribution of (i) sampling density or Grid $n$ from our lattice graph $\mathcal{F}_{lattice}$ (ii) residual connection in latticed-based graphical encoder (iii) effective proximity principle to construct graph from lattice points; for different corruption values of $P_{mask} = \{10\%, 30\%\}$.}
  
  \begin{tabular}{cccccc}
    \hline
     & Grid & Residual & Proximity & Top-1 & Top-3\\
    \hline\hline
    \multirow{8}{*}{\rotatebox{90}{Ours ($P_{mask}=10\%$)}} & \multirow{3}{*}{8} & \xmark & nearest & 26.56\% & 33.77\% \\
     & & \cmark & nearest & 36.16\% & 45.82\% \\
     & & \cmark & nearby & 38.86\% & 51.36\% \\ \cdashline{2-6}
     & \multirow{3}{*}{16} & \xmark & nearest & 14.47\% & 19.12\% \\
     & & \cmark & nearest & 40.64\% & 50.05\% \\
     & & \cmark & nearby & 42.56\% & 51.71\% \\ \cdashline{2-6}
     & \multirow{1}{*}{{32}} & \cmark & nearby & {\bf 55.50}\% & {\bf 64.72}\% \\ \cdashline{2-6}
     & \multirow{1}{*}{64} & \cmark & nearby & 45.71\% & 53.56\% \\ 
    \hline\hline
    \multirow{8}{*}{\rotatebox{90}{Ours ($P_{mask}=30\%$)}} & \multirow{3}{*}{8} & \xmark & nearest & 16.97\% & 24.02\% \\
     & & \cmark & nearest & 34.70\% & 43.76\% \\
     & & \cmark & nearby & 36.41\% & 47.75\% \\ \cdashline{2-6}
     & \multirow{3}{*}{16} & \xmark & nearest & 13.76\% & 18.43\% \\
     & & \cmark & nearest & 40.59\% & 49.56\% \\
     & & \cmark & nearby & 39.07\% & 48.41\% \\ \cdashline{2-6}
     & \multirow{1}{*}{{32}} & \cmark & nearby & {\bf 54.79}\% & {\bf 64.74}\% \\ \cdashline{2-6}
     & \multirow{1}{*}{64} & \cmark & nearby & 45.07\% & 53.50\% \\ 
     \hline
    \end{tabular}
    \label{tab:abla}
    \vspace{-0.5cm}
\end{table}

\keypoint{On Complex Sketches} {To testify the robustness and applicability of our proposed latticed representation, we further investigate the performance when dealing with complex data. Specifically, we examine the recognition accuracy of the most complicated sketches (top $25\%$) over all categories, based on the number of strokes. From the results shown in Table \ref{tab:complex}, we can see that ours outperforms other competitors when healing the most complex sketches.}

\setlength{\tabcolsep}{8pt}
\begin{table}[!h]
    \centering
    \caption{Sketch recognition accuracy on the most complex cases, i.e., top $25\%$ sketches (stroke-wise) over all categories.}
    \begin{tabular}{ccccc}
    \hline
        $P_{mask}$ & SR \cite{ha2017neural} & Sp2s \cite{chen2017sketch} & SH \cite{su2020sketchhealer} & Ours \\
        \hdashline
        $10\%$ & 0.26 & 0.18 & 0.39 & {\bf 0.43} \\
        $30\%$ & 0.03 & 0.08 & 0.37 & {\bf 0.42} \\
        \hline
    \end{tabular}
    \label{tab:complex}
\end{table}

\keypoint{Human Study} {To gain more insights about the fidelity of the healed sketches, a human study is additionally conducted. We recruited 10 participants. 50 sketch samples across all 10 classes were randomly selected. Each sample has two corrupted instances associated, at mask ratio $10\%$ and $30\%$, respectively. For each corrupted sketch, we generate a group of healed sketches using different methods (SketchHealer, SketchRNN, SketchPix2seq, and ours). We show each participant, the corrupted sketch, and the group of four healed versions in random order. Each participant is then asked to pick a healed sketch that best resembles the corrupted input. Results in Table \ref{tab:human} reveals that according to human, sketches healed by our method resemble the corrupted input the best, at both corruption levels.}

\setlength{\tabcolsep}{8pt}
\begin{table}[!h]
    \centering
    \caption{Human study on fidelity of healed sketches (in $\%$).}
    \begin{tabular}{ccccc}
    \hline
        $P_{mask}$ & SR \cite{ha2017neural} & Sp2s \cite{chen2017sketch} & SH \cite{su2020sketchhealer} & Ours \\
        \hdashline
        $10\%$ & 5.80 & 11.55 & 38.98 & {\bf 43.67}\\
        $30\%$ & 1.46 & 7.60 & 26.70 & {\bf 64.24}\\
        \hline
    \end{tabular}
    \label{tab:human}
\end{table}

\vspace{-0.3cm}
\subsection{Image-to-Sketch Synthesis}

Our Lattice-GCN-LSTM network can be applied to image-to-sketch translation. Essentially, given an input image, the corresponding edges are extracted using an off-the-shelf edge extractor \cite{zitnick2014edge}. Next, we transform the edge map into a latticed sketch representation using our lattice graph. The resulting SketchLattice could be seamlessly encoded via our graphical encoder. Finally, a vector sketch can be produced using the generative LSTM decoder. Our objective is to generate a sketch that best resembles the ground-truth sketch drawn by humans. Once trained, for any input image, we can obtain some representative lattice points based on the corresponding edges and lattice graph. Then, our generative LSTM decoder can generate a sketch from the sketch-level encoded representation $\Psi$, as stated in section~\ref{lstm}.

\keypoint{Experimental Settings} 
We use QMUL-shoe-v2 \cite{yu2016sketch}, a fine-grained sketch-based image retrieval dataset, to evaluate our image-to-sketch synthesis approach. In total, there are 6648 one-to-one mappings of image-to-sketch pairs. The dataset is split into two parts, i.e., 6000 pairs for training and the rest of 648 pairs for testing. We chose the value of $n$ for our lattice graph as $32$ and adopt the ``nearby'' strategy for graph construction. LS-SCC \cite{song2018learning}, a current state-of-the-art, is adopted for comparison.

\keypoint{Human Study} We conduct a user study that judges two aspects: (i) reality of the generated sketches, i.e., whether a sketch ``looks'' like being drawn by a human or not, and (ii) similarity between the produced sketch and its target photo. Specifically, we show triplets of images, i.e., a photo shoe, and two corresponding sketches generated by LS-SCC \cite{song2018learning} and ours in random order to 10 \textit{new} participants. Each participant is asked to, (i) choose which of the two sketches ``looks'' more like a human drawing (REAL), and (ii) identify the sketch that best resembles the photo shoe (SIM).

\begin{figure}
    \centering
    \includegraphics[width=\linewidth]{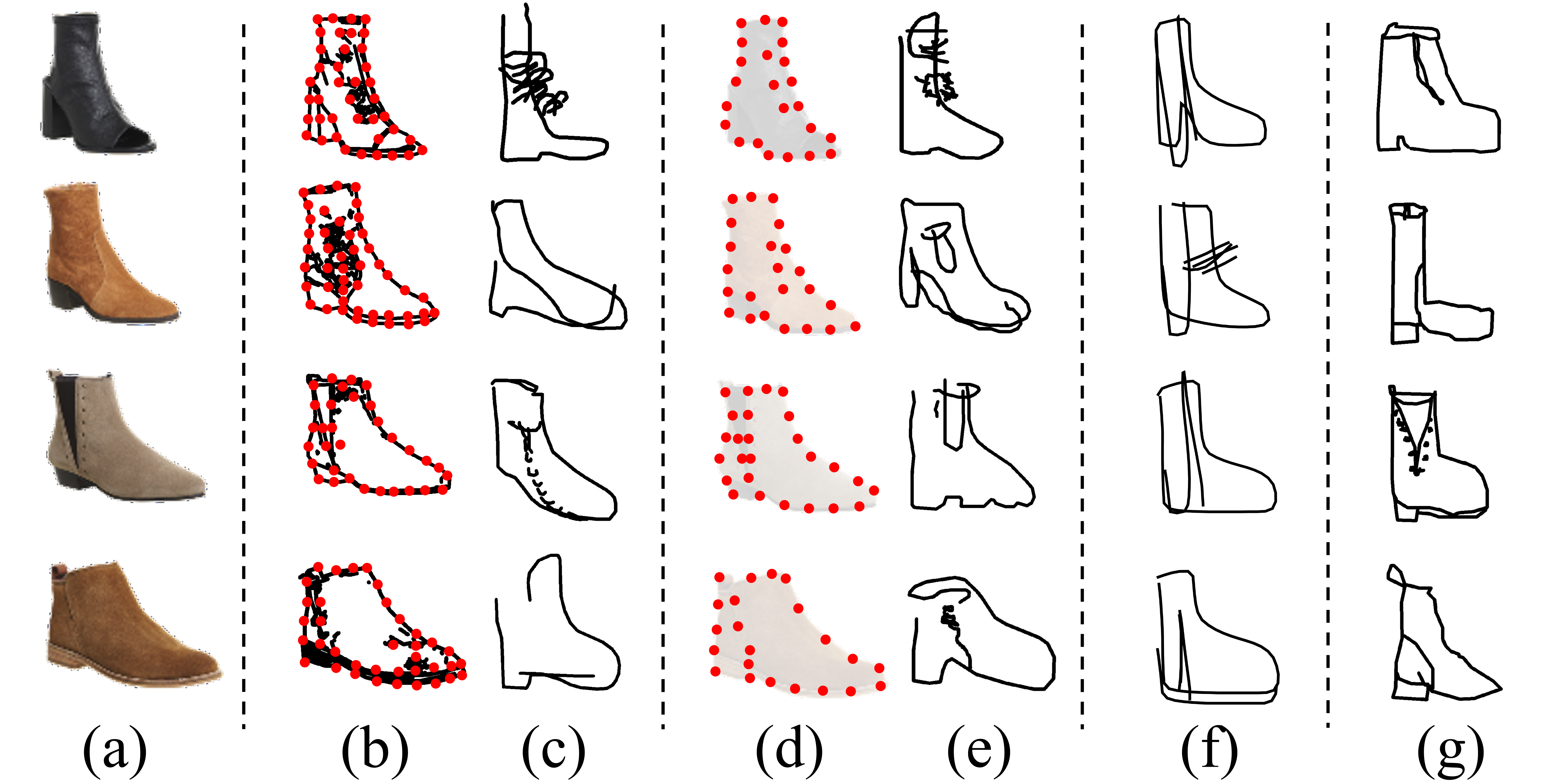}
    \caption{Image-to-sketch synthesis examples. (a) The original photos from Shoes-V2 dataset. (b) The  lattice points on edge of the photo shoes. (c) Sketches generated by our model. (d) The points introduced by human referred to the photos. (e) Sketches given by our model using lattice points shown in (d). (f) Sketches generated by LS-SCC \cite{song2018learning} for comparison. (g) Human drawn sketches according to the photos.}
    \label{fig:image-to-sketch}
    \vspace{-0.3cm}
\end{figure}

\keypoint{Results and Analysis} 
Some qualitative results are shown in Figure~\ref{fig:image-to-sketch}, where we can see that results from both LS-SCC \cite{song2018learning} and ours are far from satisfactory when compared to the human-drawn sketches, yet our generated sketches depict more detailed features, such as the ``heel'', ``sole'' and the ``zipper''. The human study results in Table~\ref{tab:image2sketch} suggest that the produced sketches by our model are closer to human drawing, while equally effective to depict real shoes compared to LS-SCC.

\begin{table}
    \centering
    \caption{Human study on reality (REAL \%) and similarity (SIM \%) of the generated sketches.}
    \begin{tabular}{ccc}
    \hline
        Method & REAL & SIM \\
        \hdashline
        LS-SCC \cite{song2018learning} &  44.82 & 49.77\\
        Ours &  {\bf 55.18} & {\bf 50.23}\\
        \hline
    \end{tabular}
    \label{tab:image2sketch}
\end{table}

\section{Conclusion}

We introduced a novel sketch representation, SketchLattice, that not only removes the bottleneck on having vector data, but also preserves the essential structural cues that vector data provides. This result in a sketch representation that is particularly amenable to structural changes that allows better abstraction modeling. We show this new representation helps multiple sketch manipulation tasks, such as sketch healing and image-to-sketch synthesis, where it outperforms state-of-the-art alternatives despite using significantly less parameters.

\vspace{-0.2cm}
\section*{Acknowledgement}
This work was supported by the National Natural Science Foundation of China (NSFC) under 61601042.

{\small
\bibliographystyle{ieee_fullname}
\bibliography{egbib}
}

\end{document}